\documentclass[prodmode,acmtiis]{acmsmall} 

\usepackage{array}
\usepackage{graphicx}
\usepackage{algorithmicx}
\usepackage{algorithm}
\usepackage{algpseudocode}
\usepackage{booktabs}
\usepackage{longtable}
\usepackage{amsmath}
\usepackage{caption}
\usepackage{sidecap}
\usepackage{enumitem}
\usepackage{adjustbox}
\usepackage{array}
\usepackage{multirow}
\usepackage{color, colortbl}
\usepackage{hhline}
\usepackage{tabularx}
\usepackage{color}

\newcolumntype{R}[2]{%
    >{\adjustbox{angle=#1,lap=\width-(#2)}\bgroup}%
    l%
    <{\egroup}%
}

\definecolor{Gray}{gray}{0.9}

\begin{document}

\markboth{Dumitrache et al.}{Crowdsourcing Ground Truth for Medical Relation Extraction}

\title{Crowdsourcing Ground Truth for Medical Relation Extraction}

\author{ANCA DUMITRACHE
\affil{Vrije Universiteit Amsterdam, CAS IBM Netherlands}
LORA AROYO
\affil{Vrije Universiteit Amsterdam}
CHRIS WELTY
\affil{Google Research}}

\begin{abstract}
Cognitive computing systems require human labeled data for evaluation, and often for training. The standard practice used in gathering this data minimizes disagreement between annotators, and we have found this results in data that fails to account for the ambiguity inherent in language. We have proposed the CrowdTruth method for collecting ground truth through crowdsourcing, that reconsiders the role of people in machine learning based on the observation that disagreement between annotators provides a useful signal for phenomena such as ambiguity in the text. We report on using this method to build an annotated data set for medical relation extraction for the $cause$ and $treat$ relations, and how this data performed in a supervised training experiment. We demonstrate that by modeling ambiguity, labeled data gathered from crowd workers can (1) reach the level of quality of domain experts for this task while reducing the cost, and (2) provide better training data at scale than distant supervision.  We further propose and validate new weighted measures for precision, recall, and F-measure, that account for ambiguity in both human and machine performance on this task.
\end{abstract}


\terms{Human Factors, Experimentation, Performance}


\acmformat{Anca Dumitrache, Lora Aroyo and Chris Welty, 2016. Crowdsourcing Ground Truth for Medical Relation Extraction.}

\begin{bottomstuff}
Author's addresses: A. Dumitrache {and} L. Aroyo, Business Web \& Media Department, Vrije Universiteit Amsterdam; C. Welty, Google Research New York.
\end{bottomstuff}

\maketitle

\section{Introduction}

Many methods for Natural Language Processing (NLP) rely on {\it gold standard} annotations, or {\it ground truth}, for the purpose of training, testing and evaluation.  Understanding {\it the role of people in machine learning} is crucial in this context, as human annotation is considered the most reliable method for collecting ground truth.  In clinical NLP and other difficult domains, researchers assume that expert knowledge of the field is required from annotators.  This means that, aside from the monetary costs of hiring humans to label data, simply finding suitable annotators bears a big time cost. The lack of annotated datasets for training and benchmarking is considered one of the big challenges of clinical NLP~\cite{chapman2011overcoming}.

Furthermore, the {\it standard data labeling practice used in supervised machine learning} often presents flaws. Data labeling is performed by humans, by reading text and following a set of guidelines to ensure a uniform understanding of the annotation task.  It is assumed that the gold standard represents a universal and reliable model for language. However, \cite{schaekermann2016} and \cite{Bayerl2011} criticize this approach by investigating the role of inter-annotator disagreement as a possible indicator of ambiguity inherent in text.  Previous experiments we performed in medical relation extraction~\cite{aroyo2013crowd} support this view by identifying two issues with the standard data labeling practice:

\begin{enumerate}

\item disagreement between annotators is usually eliminated through overly prescriptive annotation guidelines, thus creating artificial data that is neither general nor reflects the ambiguity inherent in natural language,

\item the process of acquiring ground truth by working exclusively with domain experts is costly and non-scalable, both in terms of time and money.

\end{enumerate}

Ambiguity in text also impacts automated processes for extracting ground truth.  Specifically, in the case of relation extraction from text, distant supervision~\cite{mintz2009distant,Welty:2010:LSR} is a well-established semi-supervised method that uses pairs of entities known to be related (e.g. from a knowledge base) to select sentences from a corpus that are used as positive training examples for the relations that relate the pairs. However, this approach is also prone to generating low quality training data, as not every mention of an entity pair in a sentence means a relation is also present. The problems are further compounded when dealing with ambiguous entities, or incompleteness in the knowledge base.

To address these issues, we propose the {\em CrowdTruth} method for crowdsourcing training data for machine learning. We present an alternative approach for guiding supervised machine learning systems beyond the standard data labeling practice of a universal ground truth, by instead harnessing disagreement in crowd annotations to model the ambiguity inherent in text.  We claim that, even for complex annotation tasks such as relation extraction, lack of domain expertise of the crowd is compensated by collecting a large enough set of annotations.

Previously, we studied medical relation extraction in a relatively small set of 90 sentences~\cite{aroyo2013measuring}, comparing the results from the crowd with that of two expert medical annotators. We found that disagreement within the crowd is consistent with expert inter-annotator disagreement. Furthermore, sentences that registered high disagreement tended to be vague or ambiguous when manually evaluated. In this paper, we build on these results by training a classifier for medical relation extraction with CrowdTruth data, and evaluating its performance. The goal is to show that harnessing inter-annotator disagreement results in improved performance for relation extraction classifiers. Our contributions are the following:

\begin{enumerate}

\item a comparison between using annotations from crowd and from medical experts to train a relation extraction classifier, showing that, with the processing of ambiguity, {\it classifiers trained on crowd annotations perform the same as to those trained on expert annotations};

\item a similar comparison between crowd annotations and distant supervision, showing that {\it classifiers trained on crowd annotations perform better than those trained on distant supervision};

\item a {\it dataset of 3,984 English sentences for medical relation extraction}, centering on the $cause$ and $treat$ relations, that have been processed with disagreement analysis to capture ambiguity, openly available at:

\noindent \texttt{https://github.com/CrowdTruth/Medical-Relation-Extraction}.

\end{enumerate}

\section{Related Work}

\subsection{Medical crowdsourcing}

There exists some research using crowdsourcing to collect semantic data for the medical domain. \cite{mortensen2013crowdsourcing} use crowdsourcing to verify relation hierarchies in biomedical ontologies. On 14 relations from the SNOMED CT CORE Problem List Subset, the authors report the crowd's accuracy at 85\% for identifying whether the relations were correct or not. In the field of Biomedical NLP, \cite{burger2012validating} used crowdsourcing to extract the gene-mutation relations in Medical Literature Analysis and Retrieval System Online (MEDLINE) abstracts. Focusing on a very specific gene-mutation domain, the authors report a weighted accuracy of 82\% over a corpus of 250 MEDLINE abstracts. Finally, \cite{li2015exposing} performed a study exposing ambiguities in a gold standard for drug-disease relations with crowdsourcing. They found that, over a corpus of 60 sentences, levels  of  crowd agreement varied in a similar manner to the levels of agreement  among  the  original  expert  annotators. All of these approaches present preliminary results from experiments performed with small datasets.

To our knowledge, the most extensive study of medical crowdsourcing was performed by \cite{zhai2013web}, who describe a method for crowdsourcing a ground truth for medical named entity recognition and entity linking. In a dataset of over 1,000 clinical trials, the authors show no statistically significant difference between the crowd and expert-generated gold standard for the task of extracting medications and their attributes. We extend these results by applying crowdsourcing to the more complex task of medical relation extraction, that {\it prima facie} seems to require more domain expertise than named entity recognition. Furthermore, we test the viability of the crowdsourced ground truth by training a classifier for relation extraction.

\subsection{Crowdsourcing ground truth}

Crowdsourcing ground truth has shown promising results in a variety of other domains. \cite{Snow2008} have shown that aggregating the answers of an increasing number of unskilled crowd workers with majority vote can lead to high quality NLP training data. \cite{hovy-plank-sogaard:2014:P14-2} compared the crowd versus experts for the task of part-of-speech tagging.  The authors also show that models trained based on crowdsourced annotation can perform just as well as expert-trained models. \cite{kondreddi2014combining} studied crowdsourcing for relation extraction in the general domain, comparing its efficiency to that of fully automated information extraction approaches.  Their results showed the crowd was especially suited to identifying subtle formulations of relations that do not appear frequently enough to be picked up by statistical methods.

Other research for crowdsourcing ground truth includes: entity clustering and disambiguation~\cite{Lee2013}, Twitter entity extraction~\cite{Finin2010}, multilingual entity extraction and paraphrasing~\cite{Chen2011}, and taxonomy creation~\cite{Chilton:2013}.  However, all of these approaches rely on the assumption that one black-and-white gold standard must exist for every task.  Disagreement between annotators is discarded by picking one answer that reflects some consensus, usually through using majority vote.  The number of annotators per task is also kept low, between two and five workers, in the interest of reducing cost and eliminating disagreement. \cite{NIPS2009_3644} and \cite{welinder2010multidimensional} have used a latent variable model for task difficulty, as well as latent variables to measure the skill of each annotator, to optimize crowdsourcing for image labels. The novelty in our approach is to consider language ambiguity, and consequently inter-annotator disagreement, as an inherent feature of the language. Language ambiguity can be related to, but is not necessarily a direct cause of task difficulty. The metrics we employ for determining the quality of crowd answers are specifically tailored to measure ambiguity by quantifying disagreement between annotators.

\subsection{Disagreement and ambiguity in crowdsourcing}

In addition to our own work~\cite{aroyo2013crowd}, the role of ambiguity when building a gold standard has previously been discussed by \cite{lau2014measuring}.  The authors propose a method for crowdsourcing ambiguity in the grammatical correctness of text by giving workers the possibility to pick various degrees of correctness. However, inter-annotator disagreement is not discussed as a factor in measuring this ambiguity. After empirically studying part-of-speech datasets, \cite{plank-hovy-sogaard:2014:P14-2} found that inter-annotator disagreement is consistent across domains, even across languages.  Furthermore, most disagreement is indicative of debatable cases in linguistic theory, rather than faulty annotation.  It is not unreasonable to assume that these findings manifest even more strongly for NLP tasks involving semantic ambiguity, such as relation extraction. 

In assessing the Ontology Alignment Evaluation Initiative (OAEI) benchmark, \cite{cheatham2014conference} found that disagreement between annotators (both crowd and expert) is an indicator for inherent ambiguity of alignments, and that current benchmarks in ontology alignment and evaluation are not designed to model this ambiguity. \cite{schaekermann2016} propose a framework for dealing with uncertainty in ground truth that acknowledges the notion of ambiguity, and uses disagreement in crowdsourcing for modeling this ambiguity. To our knowledge, our work presents the first experimental results of using disagreement-aware crowdsourcing for training a machine learning system.

\begin{table*}[hbt!]
\begin{center}
\caption{Set of medical relations.}
\label{tab:relation_list}
\scalebox{0.85}{
\bgroup
\def\arraystretch{1.5}
\begin{tabular}{|c|p{3.5cm}|p{4.5cm}|p{4cm}|}
\hline
{\bf Relation} & {\bf Corresponding} & {\bf Definition} & {\bf Example} \\ 
 & {\bf UMLS relation(s)} &  &   \\ \hline \hline

{\it treat} & may treat &  therapeutic use of 
a drug &  penicillin treats infection \\ \hline

{\it cause} & \parbox[t]{3.5cm}{cause of; \\ has causative agent} & the underlying reason for a symptom or a disease &   fever induces dizziness \\  \hline

{\it prevent} & may prevent &  preventative use of
a drug &  vitamin C prevents influenza \\  \hline

{\it diagnoses} & may diagnose & diagnostic use of an ingredient, test or a drug &  RINNE test is used to diagnose hearing loss \\  \hline

{\it location} & disease has primary anatomic site; has finding site  & body part 
in which disease or disorder is observed &  leukemia is found in the circulatory system \\ \hline

{\it symptom} & disease has finding; disease may have finding & deviation from normal function indicating the presence of disease or abnormality &  pain is a symptom of a broken arm \\  \hline

{\it manifestation} & has manifestation & links disorders to the observations that are closely associated with them &  abdominal distention is a manifestation of liver failure \\ \hline

{\it contraindicate} & contraindicated drug & a condition for which a drug or treatment should not be used &  patients with obesity should avoid using danazol \\  \hline

{\it side effect} & side effect & a secondary condition or symptom that results from a drug 
&  use of antidepressants causes dryness in the eyes \\  \hline

{\it associated with} & associated with & signs, symptoms or findings that often appear together &  patients who smoke often have yellow teeth \\  \hline

{\it is a} & is a & a relation that indicates that one of the terms is more specific variation of the other &  migraine is a kind of headache \\  \hline

{\it part of} & part of & an anatomical or structural sub-component &  the left ventricle is part of the heart \\  \hline
\end{tabular}
\egroup
}
\end{center}
\end{table*}

\section{Experimental Setup}

The goal of our experiments is to assess the quality of our disagreement-aware crowdsourced data in training a medical relation extraction model.  We use a binary classifier~\cite{P14-1078} that takes as input a set of sentences and two terms from the sentence, and returns a score reflecting the confidence of the model that a specific relation is expressed in the sentence between the terms.  This manifold learning classifier was one of the first to accept weighted scores for each training instance, although it still requires a discrete positive or negative label.  This property seemed to make it suitable for our experiments, as we expected the ambiguity of a sentence to impact its suitability as a training instance (in other words, we decreased the weight of training instances that exhibited ambiguity).  We investigate the performance of the classifier over two medical relations: $cause$ (between symptoms and disorders) and $treat$ (between drugs and disorders).

The quality of the crowd data in training the classifier is evaluated in two parts: first by comparing it to the performance of an expert-trained classifier, and second with a classifier trained on distant supervision data.  The training is done separately for each relation, over the same set of sentences, with different relation existence labels for crowd, expert and baseline.

\subsection{Data selection}

The dataset used in our experiments contains 3,984 medical sentences extracted from PubMed article abstracts. The sentences were sampled from the set collected by \cite{P14-1078} for training the relation extraction model that we are re-using. Wang \& Fan collected the sentences with {\em distant supervision}~\cite{mintz2009distant,Welty:2010:LSR}, a method that picks positive sentences from a corpus based on whether known arguments of the seed relation appear together in the sentence (e.g. the $treat$ relation occurs between terms $antibiotics$ and $typhus$, so find all sentences containing both and repeat this for all pairs of arguments that hold). The MetaMap parser~\cite{aronson2001effective} was used to recognize medical terms in the corpus, and the UMLS vocabulary~\cite{bodenreider2004unified} was used for mapping terms to categories, and relations to term types. The intuition of distant supervision is that since we know the terms are related, and they are in the same sentence, it is more likely that the sentence expresses a relation between them (than just any random sentence).

We started with a set of 12 relations important for clinical decision making, used also by Wang \& Fan. Each of these relations corresponds to a set of UMLS relations (Tab.\ref{tab:relation_list}), as UMLS relations are sometimes overlapping in meaning (e.g. {\it cause of} and {\it has causative agent} both map to $cause$). The UMLS relations were used as a seed in distant supervision. We focused our efforts on the relations $cause$ and $treat$. These two relations were used as a seed for distant supervision in two thirds of the sentences of our dataset (1,043 sentences for $treat$, 1,828 for $cause$). The final third of the sentences were collected using the other 10 relations as seeds, in order to make the data more heterogeneous.


To perform a comparison with expert-annotated data, we randomly sampled a set of 975 sentences from the distant supervision dataset. This set restriction was done not just due to the cost of the experts, but primarily because of their limited time and availability. To collect this data, we employed medical students, in their third year at American universities, that had just taken  United States Medical Licensing Examination (USMLE) and were waiting for their results.  Each sentence was annotated by exactly one person. The annotation task consisted of deciding whether or not the UMLS seed relation discovered by distant supervision is present in the sentence for the two selected terms. The expert annotation costs are about \$2.00 per sentence.

The crowdsourced annotation setup is based on our previous medical relation extraction work~\cite{aroyo2014threesides}. For every sentence, the crowd was asked to decide which relations (from Tab.\ref{tab:relation_list}) hold between the two extracted terms. The task was multiple choice, workers being able to choose more than one relation at the same time.  There were also options available for cases when the medical relation was other than the ones we provided ($other$), and for when there was no relation between the terms ($none$). The crowdsourcing was run on the CrowdFlower\footnote{\texttt{https://crowdflower.com/}} platform, with 15 workers per sentence, at a cost of \$0.66 per sentence. Compared to a single expert judgment, the cost per sentence of the crowd amounted to  $2/3$ of the sum paid for the experts.

All of the data that we have used, together with the templates for the crowdsourcing tasks, and the crowdsourcing implementation details are available online at:

\noindent \texttt{https://github.com/CrowdTruth/Medical-Relation-Extraction}.

\subsection{CrowdTruth metrics}

The crowd output was processed with the use of CrowdTruth metrics -- a set of general-purpose crowdsourcing metrics~\cite{inel2014crowdtruth}, that have been successfully used to model ambiguity in annotations for relation extraction, event extraction, sounds, images, and videos~\cite{aroyo2014threesides}.  These metrics model ambiguity in semantic interpretation based on the triangle of reference~\cite{Ogden1923}, with the vertices being the input sentence, the worker, and the seed relation.  Ambiguity and disagreement at any of the vertices (e.g. a sentence with unclear meaning, a poor quality worker, or an unclear relation) will propagate in the system, influencing the other components.  For example, if a sentence is unclear, we expect workers will be more likely to disagree with each other; if a worker is not doing a good job, we expect that worker to disagree with other workers across the majority of the sentences they worked on; and if a particular target relation is unclear, we expect workers to disagree on the application of that relation across all the sentences. By using multiple workers per sentence and requiring each worker to annotate multiple sentences, the aggregate data helps us isolate these individual signals and how they interact.  Thus a high quality worker who annotates a low clarity sentence will be recognized as high quality. In our workflow, these metrics are used both to eliminate spammers, as detailed by~\cite{aroyo2014threesides}, and to determine the clarity of the sentences and relations. The main concepts are:

\begin{itemize}

\item {\it annotation vector:} used to model the annotations of one worker for one sentence.  For each worker $i$ submitting their solution to a task on a sentence $s$, the vector $W_{s,i}$ records their answers. If the worker selects a relation, its corresponding component would be marked with `1', and `0' otherwise.  The vector has 14 components, one for each relation, as well as $none$ and $other$. Multiple choices (e.g. picking multiple relations for the same sentence) are modeled by marking all corresponding vector components with `1'.

\item {\it sentence vector:} the main component for modeling disagreement.  For every sentence $s$, it is computed by adding the annotation vectors for all workers on the given task: $V_{s} = \sum_{i}{W_{s,i}}$ .  One such vector was calculated for every sentence.   

\item {\it sentence-relation score:} measures the ambiguity of a specific relation in a sentence with the use of cosine similarity. The higher the score, the more clearly the relation is expressed in the sentence. The sentence-relation score is computed as the cosine similarity between the sentence vector and the unit vector for the relation: $ srs(s, r) = cos(V_s, \hat{r}) $, where the unit vector $\hat{r}$ refers to a vector where the component corresponding to relation $r$ is equal to `1', and all other components are equal to `0'. The reasoning is that the unit vector $\hat{r}$ corresponds to the clearest representation of a relation in a sentence -- i.e. when all workers agree that relation $r$ exists between the seed terms, and all other relations do not exist. As a cosine similarity, these scores are in the $[0, 1]$ interval.  Tab.\ref{tab:scores} shows the transformation of sentence vectors to the sentence-relation scores and then to the training scores using the threshold below.

\item {\it sentence-relation score threshold:} a fixed value in the interval $[0, 1]$ used to differentiate between a negative and a positive label for a relation in a sentence. Given a value $t$ for the threshold, all sentences with a sentence-relation score less than $t$ get a negative label, and the ones with a score greater or equal to $t$ are positive. The results section compares the performance of the crowd at different threshold values. This threshold was necessary because our classifier required either a positive or negative label for each training example. Therefore, the sentence-relation scores must be re-scaled in the $[-1, 0]$ interval for negative labels. An example of how the crowd scores for training the model were calculated is given in Tab.\ref{tab:scores}.

\end{itemize}

\newcolumntype{g}{>{\columncolor{Gray}}c}
\begin{table}[h!]

\caption{Given two sentences, {\it Sent.1} and {\it Sent.2}, with term pairs in bold font, the table shows the transformation of the sentence vectors to sentence -- relation scores, and then to {\it crowd} scores used for model training. The sentence-relation threshold for the train score is set at $0.5$ for these examples.

{\it Sent.1:} {\bf Renal osteodystrophy} is a general complication of chronic renal failure and {\bf end stage renal disease}.

{\it Sent.2:} If {\bf TB} is a concern, a {\bf PPD} is performed.}
\label{tab:scores}

\centering
\begin{tabular}{|r|cc|cc|cc|}
\hline
\multirow{3}{*}{{\bf Relation}} & \multicolumn{2}{c|}{{\bf sentence}} & \multicolumn{2}{c|}{{\bf sentence -- relation}} & \multicolumn{2}{c|}{{\bf crowd score used in}} \\
& \multicolumn{2}{c|}{{\bf vector}} & \multicolumn{2}{c|}{{\bf score}} & \multicolumn{2}{c|}{{\bf model training}} \\ \cline{2-7}
& {\it Sent.1} & {\it Sent.2} & {\it Sent.1} & {\it Sent.2} & {\it Sent.1} & {\it Sent.2} \\ \hline
{\it treat}           & 0  & 3 & 0    & 0.36 & -1    & -0.64 \\
\cellcolor{Gray}{\it prevent}         & \cellcolor{Gray}0  & \cellcolor{Gray}1 & \cellcolor{Gray}0    & \cellcolor{Gray}0.12 & \cellcolor{Gray}-1    & \cellcolor{Gray}-0.88 \\
{\it diagnose}        & 1  & 7 & 0.09 & 0.84 & -0.91 & 0.84 \\
\cellcolor{Gray}{\it cause}           & \cellcolor{Gray}10 & \cellcolor{Gray}0 & \cellcolor{Gray}0.96 & \cellcolor{Gray}0    & \cellcolor{Gray}0.96  & \cellcolor{Gray}-1 \\
{\it location}        & 1  & 0 & 0.09 & 0    & -0.91 & -1 \\
\cellcolor{Gray}{\it symptom}         & \cellcolor{Gray}2  & \cellcolor{Gray}0 & \cellcolor{Gray}0.19 & \cellcolor{Gray}0    & \cellcolor{Gray}-0.81 & \cellcolor{Gray}-1 \\
{\it manifestation}   & 0  & 0 & 0    & 0    & -1    & -1 \\
\cellcolor{Gray}{\it contraindicate}  & \cellcolor{Gray}0  & \cellcolor{Gray}0 & \cellcolor{Gray}0    & \cellcolor{Gray}0    & \cellcolor{Gray}-1    & \cellcolor{Gray}-1 \\
{\it associated with} & 1  & 3 & 0.09 & 0.36 & -0.91 & -0.64 \\
\cellcolor{Gray}{\it side effect}     & \cellcolor{Gray}0  & \cellcolor{Gray}0 & \cellcolor{Gray}0    & \cellcolor{Gray}0    & \cellcolor{Gray}-1    & \cellcolor{Gray}-1 \\
{\it is a}            & 0  & 0 & 0    & 0    & -1    & -1 \\
\cellcolor{Gray}{\it part of}         & \cellcolor{Gray}0  & \cellcolor{Gray}0 & \cellcolor{Gray}0    & \cellcolor{Gray}0    & \cellcolor{Gray}-1    & \cellcolor{Gray}-1 \\
{\it other}           & 0  & 1 & 0    & 0.12 & -1    & -0.88 \\
\cellcolor{Gray}{\it none}            & \cellcolor{Gray}0  & \cellcolor{Gray}0 & \cellcolor{Gray}0    & \cellcolor{Gray}0    & \cellcolor{Gray}-1    & \cellcolor{Gray}-1 \\ \hline
\end{tabular}
\end{table}

\subsection{Training the model}

The sentences together with the relation annotations were then used to train a manifold model for relation extraction~\cite{P14-1078}.  This model was developed for the medical domain, and tested for the relation set that we employ.  It is trained per individual relation, by feeding it both {\it positive} and {\it negative} data.  It offers support for both discrete labels, and real values for weighting the confidence of the training data entries, with positive values in $(0, 1]$, and negative values in $[-1, 0)$. Using this system, we train several models using five-fold cross validation, in order to assess the performance of the crowd dataset. The training was done separately for the $treat$ and $cause$ relations. For each relation, we constructed four datasets, with the same sentences and term pairs, but with different labels for whether or not the relation is present in the sentence:

\begin{enumerate}

\item {\it baseline:} The distant supervision data is used to provide discrete (positive or negative) labels on each sentence - i.e. if a sentence contains two terms known (in UMLS) to be related by {\it treats}, the sentence is considered positive. Distant supervision does not extract negative examples, so in order to generate a negative set for one relation, we use positive examples for the other (non-overlapping) relations shown in Tab.~\ref{tab:relation_list}. This dataset constitutes the baseline against which all other datasets are tested.

\item {\it expert:} Discrete labels based on an expert's judgment as to whether the {\it baseline} label is correct. The experts do not generate judgments for all combinations of sentences and relations -- for each sentence, the annotator decides on the seed relation extracted with distant supervision.  Similarly to the baseline data, we reuse positive examples from the other relations to increase the number of negative examples.

\item {\it single:} Discrete labels for every sentence are taken from one randomly selected crowd worker who annotated the sentence.  This data simulates the traditional single annotator setting common in annotation environments.

\item {\it crowd:} Weighted labels for every sentence are based on the CrowdTruth {\it sentence-relation score}.  Labels are separated into a positive and negative set based on the {\it sentence-relation score threshold}, and negative labels are rescaled in the $[-1, 0]$ interval. An example of how the scores were processed is given in Tab.\ref{tab:scores}.

\end{enumerate}

For each relation, two experiments were run. First, we performed a comparison between the {\it crowd} and {\it expert} datasets by training a model using the subset of sentences that also has expert annotations. In total there are 975 unique sentences in this set. After we were able to determine the quality of the {\it crowd} data, we performed a second experiment comparing the performance of the classifier when trained with the {\it crowd} and {\it baseline} annotations from the full set of 3,984 sentences.


\subsection{Evaluation data}

In order for a meaningful comparison between the crowd and expert models, the evaluation set needs to be carefully vetted. For each of the relations, we started by selecting the positive/negative threshold for {\it sentence-relation score} such that the crowd agrees the most with the experts. We assume that, if both the expert and the crowd agree that a sentence is either a positive or negative example, it can automatically be used as part of the test set. Such a sentence was labeled with the expert score.


The interesting cases appear when crowd and expert disagree. To ensure a fair comparison, our team adjudicated each of them to decide whether or not the relation is present in the sentence. The sentences where no decision could be reached were subsequently removed from the evaluation.  There were 32 such sentences for $cause$ (18 with negative expert labels, and 14 with positive), and 15 for $treat$ (all for positive expert labels).  Table~\ref{tab:bad_sent} in the Appendix shows some example sentences that were removed from the evaluation set.  This set constitutes of confusing and ambiguous sentences that our team could not agree on.  Often these sentences contained a vague association between the two terms, but the relation was too broad to label it as a positive classification example.  However, because a relation is nevertheless present, these sentences cannot be labeled as negative examples either.  Eliminating these sentences is a disadvantage to a system like ours which was motivated specifically by the need to handle such cases, however the scientific community still only recognizes discrete measures such as precision and recall, and we felt it only fair to eliminate the cases where we could not agree on the correct way to map ambiguity into a discrete score.

For evaluation, we selected sentences through 5-fold cross-validation, but we obviously only used the test labels when a partition was chosen to be test. For the second evaluation over 3,984 sentences, we again selected test sets using cross-validation over the sentences with expert annotation, adding the unselected sentences with their training labels to the training set. This allows us to directly compare the learning curves between the 975 and 3,984 sentences experiments. The scores reported are the mean over the cross-validation runs.

\subsection{CrowdTruth-weighted evaluation}

We also explored how to incorporate CrowdTruth into the evaluation process. The reasoning of our approach is that the ambiguity of a sentence should also be accounted for in the evaluation -- i.e. sentences that do not clearly express a relation should not count for as much as clear sentences. In this case, the {\it sentence-relation score} gives a real-valued score that measures the degree to which a particular sentence expresses a particular relation between two terms. Therefore, we propose a set of evaluation metrics that have been weighted with the {\it sentence-relation score} for a given relation. The metrics have been previously tested on a subset of our ground truth data, as detailed in~\cite{dumitrache2015a}.

We collect true and false positives and negatives in the standard way, such that $tp(s) = 1 $ iff $s$ is a true positive, and $0$ otherwise, similarly for $fp, tn, fn$. The positive sentences (i.e true positive and false negative labels) are weighted with the sentence-relation score $srs(s)$ for the given sentence-relation pair, i.e. how likely it is that the relation is expressed in the sentence. Negative sentences (true negative and false positive labels) are weighted with $1 - srs(s)$, how likely it is that that the sentence does not express the relation. Based on this, we define the following metrics to be used in the evaluation:

\begin{itemize}

\item {\it weighted precision:}   Where normally $P = tp/(tp + fp)$, weighted precision $$P' = \dfrac{\sum_{s}{srs(s) \cdot tp(s)}} {\sum_{s}{srs(s) \cdot tp(s) + (1 - srs(s)) \cdot fp(s)}};$$

\item {\it weighted recall:} Where normally $R = tp/(tp + fn)$, weighted recall $$R' = \dfrac{\sum_{s}{srs(s) \cdot tp(s)}}{\sum_{s}{srs(s) \cdot tp(s) + srs(s) \cdot fn(s)}};$$

\item {\it weighted F-measure:} Is the harmonic mean of weighted precision and recall: $$F1' = 2 P' R' / (P' + R').$$

\end{itemize}

\section{Results}

\subsection{CrowdTruth vs. medical experts}

\begin{figure}[htb!]
\centering
\begin{minipage}{.44\textwidth}
\centering
\includegraphics[width=\textwidth]{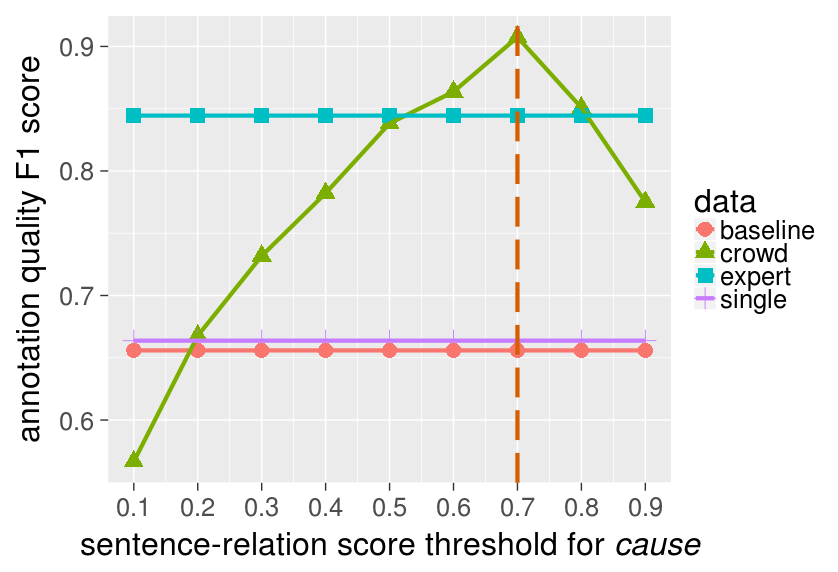}
\caption{Annotation quality F1 scores for the $cause$ relation.}
\label{fig:cause_ann_quality}
\end{minipage} \qquad 
\begin{minipage}{.44\textwidth}
\centering
\includegraphics[width=\textwidth]{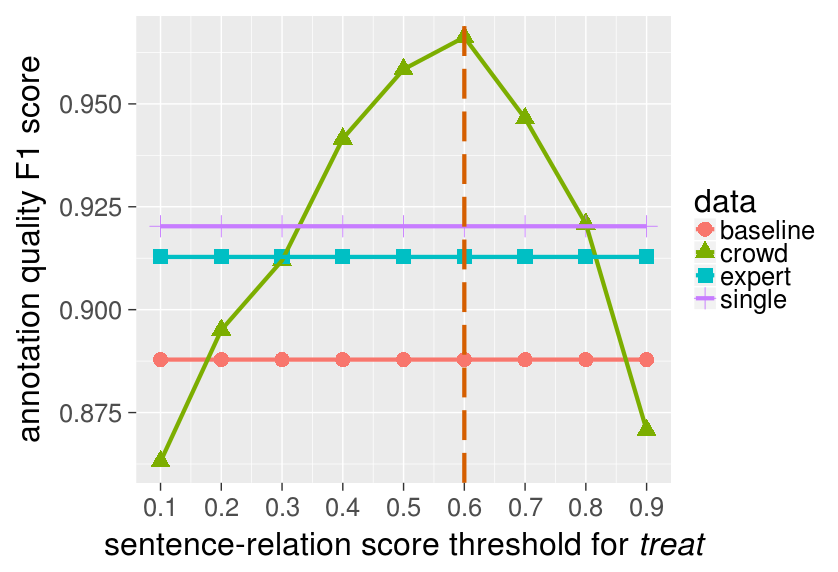}
\caption{Annotation quality F1 scores for the $treat$ relation.}
\label{fig:treat_ann_quality}
\end{minipage}
\end{figure}
\begin{figure}[htb!]
\centering
\begin{minipage}{.44\textwidth}
\centering
\includegraphics[width=\textwidth]{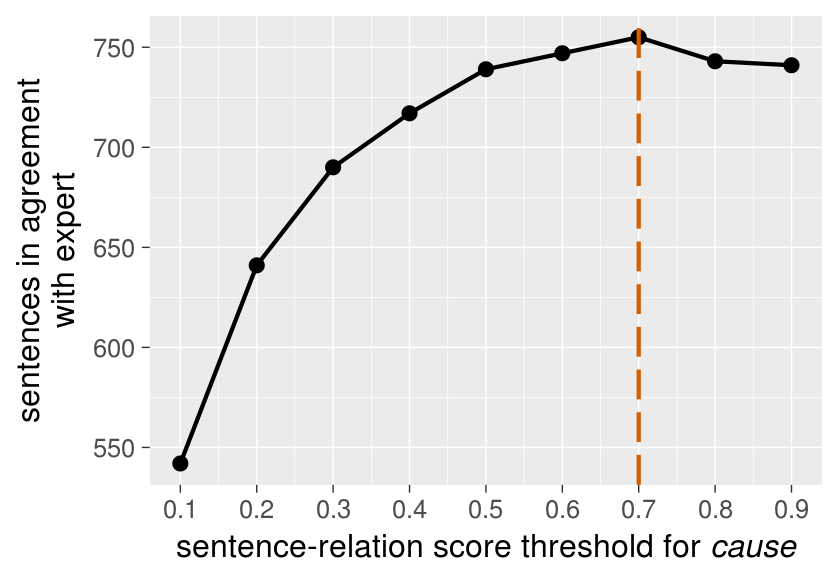}
\caption{Crowd \& expert agreement for $cause$ relation.}
\label{fig:cause_crowd_exp_agr}
\end{minipage} \qquad
\begin{minipage}{.44\textwidth}
\centering
\includegraphics[width=\textwidth]{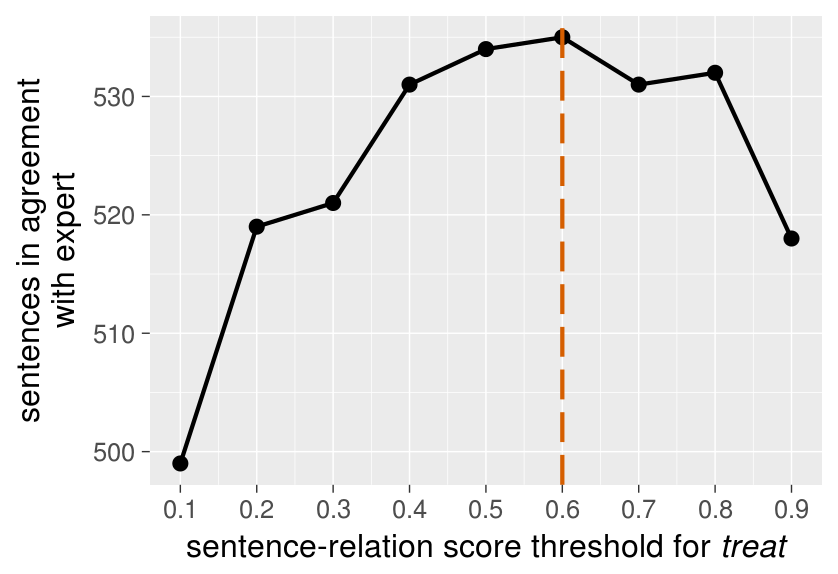}
\caption{Crowd \& expert agreement for $treat$ relation.}
\label{fig:treat_crowd_exp_agr}
\end{minipage}
\end{figure}

In the first experiment, we compare the quality of the crowd with expert annotations over the sentences that have been also annotated by experts.  We start by comparing the crowd and expert labels to the adjudicated test labels on each sentence, without training a classifier, computing an F1 score that measures the {\em annotation quality} of each set, shown in Fig.\ref{fig:cause_ann_quality} \& \ref{fig:treat_ann_quality}. Since the baseline, expert, and single sets are binary decisions, they appear as horizontal lines, whereas the crowd annotations are shown at different sentence-relation score thresholds. For both relations, the crowd labels have the highest annotation quality F1 scores, 0.907 for the $cause$ relation, and 0.966 for $treat$. The expert data is close behind, with an F1 score of 0.844 for $cause$ and 0.912 for $treat$. To calculate the statistical significance of the results, we used McNemar's test~\cite{mcnemar1947note} over paired nominal data, by constructing a contingency table from the binary classification results (i.e. correct or incorrect classification) of paired datasets (e.g. crowd and expert). This difference between crowd and expert is not significant for $cause$ ($p > 0.5$, $\chi^2 = 0.034$), and significant for $treat$ ($p = 0.002$, $\chi^2 = 5.127$). The sentence -- relation score threshold for the best annotation quality F1 is also the threshold where the highest agreement between crowd and expert occurs (Fig.\ref{fig:cause_crowd_exp_agr} \& \ref{fig:treat_crowd_exp_agr}).

\begin{figure}[htb!]
\centering
\begin{minipage}{.44\textwidth}
\centering
\includegraphics[width=\textwidth]{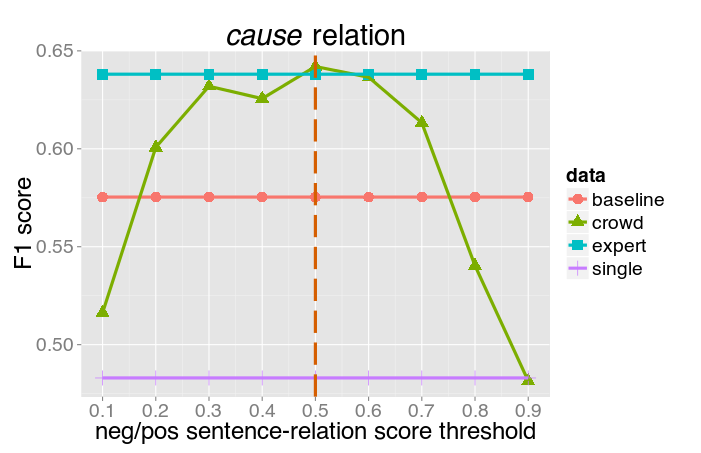}
\caption{Model testing F1 scores for the $cause$ relation.}
\label{fig:cause_exp_f1}
\end{minipage} \qquad 
\begin{minipage}{.44\textwidth}
\centering
\includegraphics[width=\textwidth]{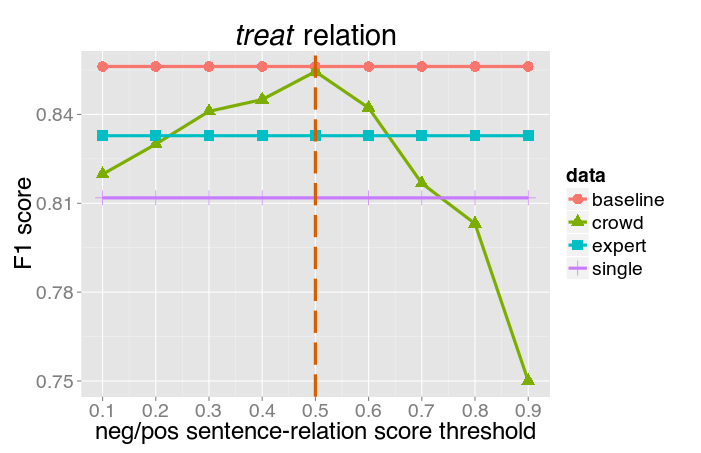}
\caption{Model testing F1 scores for the $treat$ relation.}
\label{fig:treats_exp_f1}
\end{minipage}
\end{figure}

Next we compare the quality of the crowd and expert annotations by training the relation extraction model using each dataset. For the $cause$ relation, the results of the evaluation (Fig.\ref{fig:cause_exp_f1}) show the best performance for the crowd-trained model when the sentence-relation threshold is 0.5. Trained with this data, the classifier model achieves an F1 score of 0.642, compared to the expert-trained model which reaches 0.638. The difference is statistically significant with $p = 0.016$ ($\chi^2 = 5.789$).

Tab.\ref{tab:statistic_exp} shows the full results of the evaluation, together with the results of the CrowdTruth weighted metrics (P', R', F1'). In all cases, the F1' score is greater than F1, indicating that ambiguous sentences have a strong impact on the performance of the classifier. Weighted P' and R' also have higher values in comparison with simple precision and recall.

\begin{table}[htb!]
\centering
\caption{Model evaluation results over sentences with expert annotation. Crowd scores are shown at 0.5 negative/positive sentence-relation score threshold.}
\label{tab:statistic_exp}
\begin{tabular}{|r|c|cc|cc|cc|}

\hline
& {\bf Dataset} & {\bf P} & {\bf P'} & {\bf R} & {\bf R'} & {\bf F1} & {\bf F1'}   \\ \hline 
& {\it crowd} & 0.565 & 0.632 & 0.743 & 0.754 & {\bf 0.642} & {\bf 0.687} \\ 
{\it cause} & {\it expert} & {\bf 0.672} & {\bf 0.711} & 0.604 & 0.616 & 0.638 & 0.658 \\ 
relation & {\it baseline} & 0.436 & 0.474 & {\bf 0.844} & {\bf 0.842} & 0.575 & 0.606 \\ 
& {\it single} & 0.495 & 0.545 & 0.473 & 0.478 & 0.483 & 0.658 \\ \hline 
& {\it crowd} & 0.823 & 0.843 & 0.891 & 0.902 & 0.854 & 0.869 \\ 
{\it treat} & {\it expert} & {\bf 0.834} & {\bf 0.863} & 0.833 & 0.84 & 0.832 & 0.85 \\ 
relation & {\it baseline} & 0.767 & 0.811 & {\bf 0.968} & {\bf 0.968} & {\bf 0.856} & {\bf 0.882} \\ 
& {\it single} & 0.774 & 0.819 & 0.856 & 0.866 & 0.811 & 0.84 \\ \hline 
\end{tabular}
\end{table}

For the $treat$ relation, the results of the evaluation (Fig.\ref{fig:treats_exp_f1}) shows baseline as having the best performance, at an F1 score of 0.856. The crowd dataset, with an F1 score of 0.854, still out-performs the expert, scoring at 0.832.  These three scores are not, however, significantly different ($p > 0.5$, $ \chi^2 = 0.453 $), as there are so few actual pairwise differences (a consequence of the higher scores and the size of the dataset). 

For both $cause$ and $treat$ relations, the single annotator dataset performed the worst. It is also worth noting that the sentence -- relation score threshold for the best classifier performance (0.5 for both relations) is different from the threshold for best annotation quality, and highest agreement with expert (0.7 for $cause$ and 0.6 for $treat$, Fig.\ref{fig:cause_ann_quality} \& \ref{fig:treat_ann_quality}).

\begin{figure}[htb!]
\begin{minipage}{0.45\textwidth}
\centering
\includegraphics[width=\textwidth]{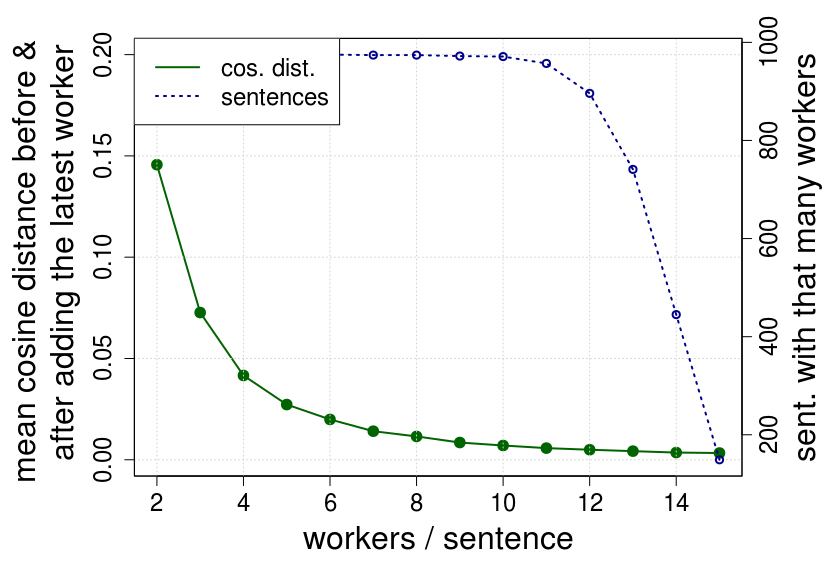}
\caption{Mean cosine distance for sentence vectors before and after adding the latest worker, shown per number of workers.}
\label{fig:cos_sim_per_worker}
\end{minipage} \qquad
\begin{minipage}{0.45\textwidth}
\centering
\includegraphics[width=\textwidth]{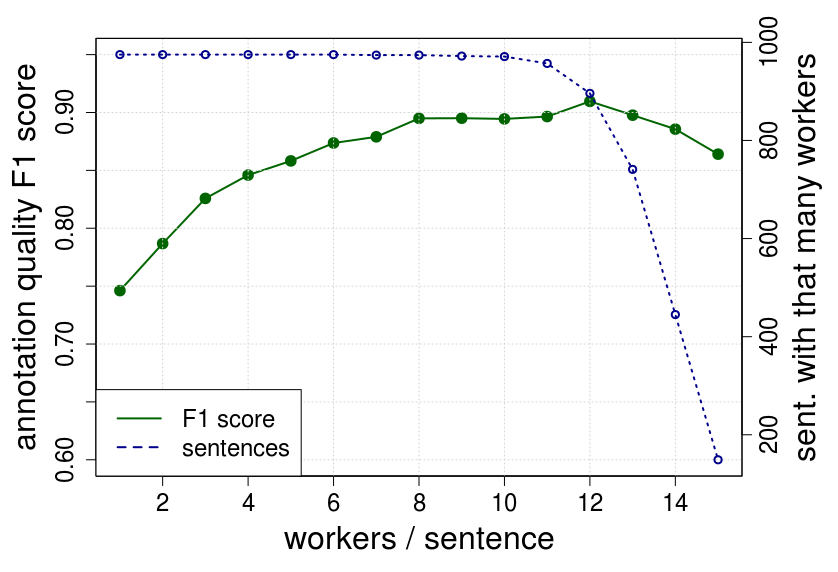}
\caption{Combined annotation quality F1 for $cause$ and $treat$ crowd, at their best pos./neg. thresholds (Fig.\ref{fig:cause_ann_quality}\&\ref{fig:treat_ann_quality}), per number of workers.}
\label{fig:ann_quality_f1}
\end{minipage}
\end{figure}

Finally, we checked whether the number of workers per task was sufficient to produce a stable sentence-relation score. We did this in two ways, first by measuring the cosine distance between the sentence vectors at each incremental number of workers (Fig.~\ref{fig:cos_sim_per_worker}), and second by measuring the annotation quality F1 score for $treat$ and $cause$, combined using the micro-averaged method (i.e. adding up the individual true positives, false positives etc.), against the number of workers annotating each sentence (Fig.~\ref{fig:ann_quality_f1}). For both plots, the workers were added in the order that they submitted their results on the crowdsourcing platform.  Based on these results, we decided to ensure that each sentence has been annotated by at least 10 workers after spam removal.  The plot of the mean cosine distance between sentence vectors before and after adding the latest worker shows that the sentence vector is stable at 10 workers. The annotation quality F1 score per total number of workers plot appears less stable in general, with a peak at 12 workers, and a subsequent drop due to sparse data -- only 149 sentences had 15 or more total workers. However, after 10 workers there are no significant increases in the annotation quality.  While it can be argued that both plots stabilize for a lower number of workers, we picked 10 as a threshold because it gives some room for improvement for sentences that might need more workers before getting a stable score, while still being economical.

\subsection{CrowdTruth vs. distant supervision}

Distant supervision is a widely used technique in NLP, because its obvious flaws can be overcome at scale.  We did not have enough time with the experts to gather a larger dataset from them, but the crowd is always available, so after we determined that the performance of the crowd matched the medical experts, we extended the experiments to 3,984 sentences. The crowd dataset in this experiment uses a fixed sentence-relation score threshold equal to 0.5, since this is the value where the crowd performed the best in the previous experiment, for both of the relations. As in the previous experiment, we employed five-fold cross validation to train the model. The test sets were kept the same as in the previous experiment, using the test partition labels as a gold standard. The goal was to compare the crowd to the distant supervision baseline, while scaling the number of training examples, until achieving a stable learning curve in the F1 score. Since the single annotator dataset performed badly in the initial experiment, it was dropped from this analysis. The full results of the experiment are available in Tab.\ref{tab:statistic_bas}.

\begin{figure}[htb!]
\centering
\begin{minipage}{.44\textwidth}
\centering
\includegraphics[width=\textwidth]{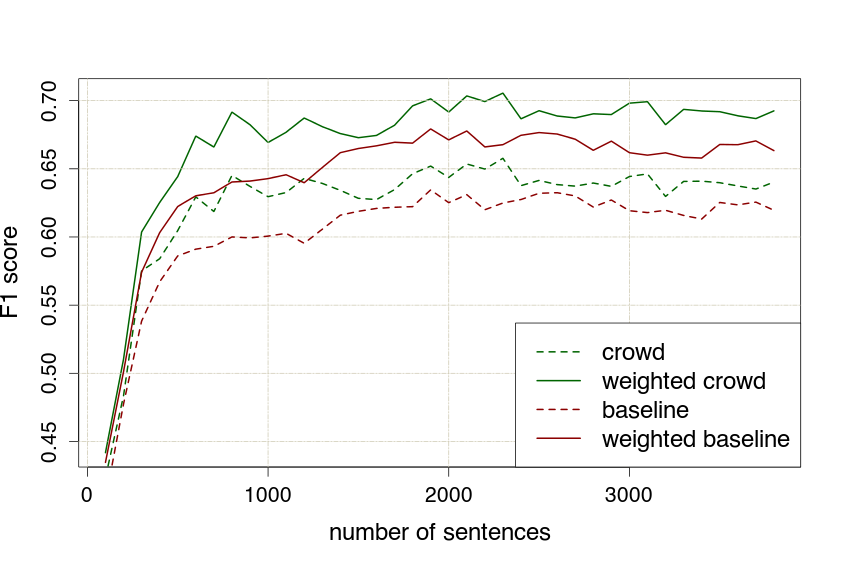}
\caption{Learning curves for $cause$ relation.}
\label{fig:cause_3800}
\end{minipage} \qquad
\begin{minipage}{.44\textwidth}
\centering
\includegraphics[width=\textwidth]{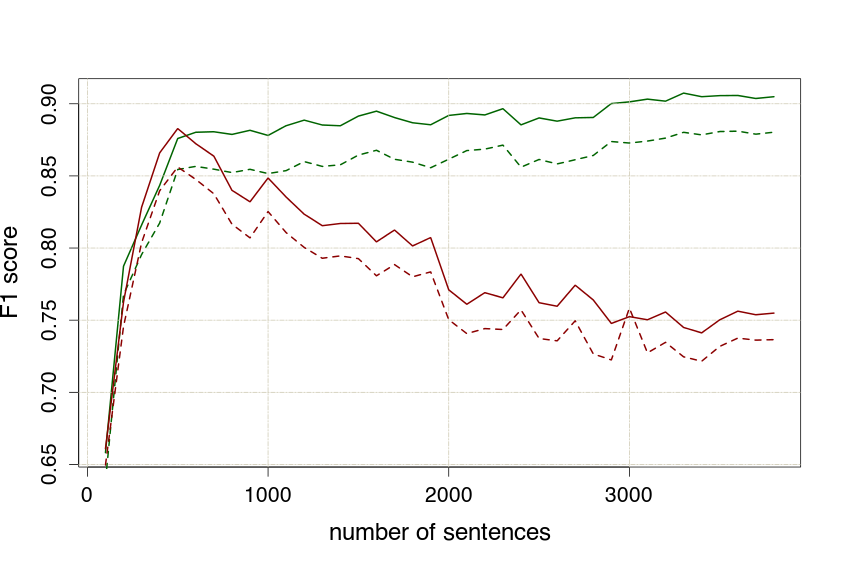}
\caption{Learning curves for $treat$ relation.}
\label{fig:treat_3800}
\end{minipage}
\end{figure}

\begin{table}[!htb]
\centering
\caption{Model evaluation results over 3,984 sentences. Crowd scores are shown at 0.5 sentence-relation score threshold.}
\label{tab:statistic_bas}
\begin{tabular}{|r|c|cc|cc|cc|}
\hline
& {\bf Dataset} & {\bf P} & {\bf P'} & {\bf R} & {\bf R'} & {\bf F1} & {\bf F1'}   \\ \hline
{\it cause} & {\it crowd} & {\bf 0.538} & {\bf 0.61} & 0.79  & 0.802 & {\bf 0.64}  & {\bf 0.692} \\ 
relation & {\it baseline} & 0.475 & 0.53 & {\bf 0.889} & {\bf 0.887} & 0.619 & 0.663 \\ \hline 
{\it treat} & {\it crowd} & {\bf 0.876} & {\bf 0.913} & {\bf 0.887} & {\bf 0.898} & {\bf 0.88} & {\bf 0.904} \\ 
relation & {\it baseline} & 0.808 & 0.858 & 0.678 & 0.673 & 0.736 & 0.754 \\ \hline 
\end{tabular}
\end{table}

For both relations, the crowd consistently performs better than the baseline. In the case of the $cause$ relation, crowd and baseline perform closer to each other, with an F1 score of 0.64 for crowd and 0.619 for baseline. This difference is significant with $ p = 0.001 $ and $\chi^2 = 10.028$. The gap in performance is even greater for accuracy, where the  crowd model scored at 0.773 and baseline at 0.705. The learning curves for the $cause$ relation (Fig.\ref{fig:cause_3800}) show both datasets achieve stable performance.

For the $treat$ relation, the crowd scores an F1 of 0.88, while baseline scores 0.736, with $p = 1.39 \times 10^{-10} $ significance, and $\chi^2 = 41.176$. The learning curves (Fig.\ref{fig:treat_3800}) show that, while baseline out-performed crowd when training with less than 1,000 sentences, crowd performance became stable after 1,000, while baseline went down, significantly increasing the gap between the two datasets.

The gap in performance is also present in the weighted F1' metrics. As is the case in the previous experiment, the F1' scores higher than the regular F1 score for both crowd and baseline. The only weighted metric that does not increase is the baseline recall. This is also the only metric by which the baseline model performed better than the crowd.

\section{Discussion}

\subsection{CrowdTruth vs. medical experts}

Our first goal was to demonstrate that, like the crowdsourced medical entity recognition work by \cite{zhai2013web}, the CrowdTruth approach of having multiple annotators with precise quality scores can be harnessed to create gold standard data with a quality that rivals annotated data created by medical experts.  Our results show this clearly, in fact with slight improvements, with a sizable dataset (975 sentences) on a problem (relation extraction) that {\em prima facie} seems to require more domain expertise (than entity recognition).

The most interesting result of the first experiment is that the sentence-relation score threshold that gives the best F1 score is the same for both $cause$ (Fig.\ref{fig:cause_exp_f1}) and $treat$ (Fig.\ref{fig:treats_exp_f1}) relations, at a value of 0.5. This shows that ambiguous data is indeed valuable in training of clinical NLP models, and that being too strict with what constitutes a positive (or negative) training example produces flawed ground truth data. It is also worth noting that the single crowd annotator performs the worst for each of the relations. This could be further indication that the crowd can only achieve quality when accounting for the choices of multiple annotators, and further calls into question the standard practice of using only one annotator per example.

A curious aspect of the results is that the sentence-relation score threshold that gives the highest annotation quality F1 score (i.e. F1 score calculated directly over the test data, without training the model), shown in Fig.\ref{fig:cause_ann_quality} \& \ref{fig:treat_ann_quality}, is different from the best threshold for classifier performance (Fig.\ref{fig:cause_exp_f1} \& \ref{fig:treats_exp_f1}). It is the lower threshold (equal to 0.5) that results in the best model. This is most likely due to the higher recall of the lower threshold, which exposes the classifier to more positive examples.  F-score is the harmonic mean between precision and recall, and does not necessarily represent the best trade-off between them, as this experiment shows for annotation quality.  Indeed F-score may not be the best trade-off between precision and recall for the classifier, either, but it is the most widely accepted and reported metric for relation extraction.  Note also that for both relations, the annotation quality at the 0.5 threshold is comparable or better than expert annotation quality.


The fact that the experts performed slightly worse than the single crowd annotator on the $treat$ annotation quality (Fig.\ref{fig:treat_ann_quality}) is also a surprising finding. Although the difference is too small to draw significant conclusions from, it indicates that the $treat$ relation was easier to interpret by the crowd and generated less disagreement -- the single annotator had a better performance for $treat$ than for $cause$ also in the model evaluation (Fig.\ref{fig:treats_exp_f1}). This result also shows that the experts we employed were fallible, and made mistakes when annotating the data. A better approach to gather the expert annotations would be to ask several experts per sentence, to account for the failures in a single person's interpretations.

In our error analysis of the annotation quality, we found that (as Figs. \ref{fig:cause_ann_quality} \& \ref{fig:treat_ann_quality} show) experts and the crowd both make errors, but of different kinds.  Experts tend to see relations that they know hold as being expressed in sentences, when they are not.  For example, in, ``He was the first to describe the relation between {\bf Hemophelia} and {\bf Hemophilic Arthropathy},'' experts labeled the sentence as expressing the $cause$ relation, since they know Hemophelia causes Hemophilic Arthropathy.  Thus they are particularly prone to errors in sentences selected by distant supervision, since that is the selection criterion. Table~\ref{tab:bad_expert} from the Appendix shows more such examples.  Crowd workers, on the other hand, were more easily fooled by sentences that expressed one of the target relations, but {\em not between the selected arguments.}  For example, in ``Influenza treatments such as {\bf antivirals} and {\bf antibiotics} are sometimes recommended," some crowd workers will label the sentence with treats, even though we are looking for the relation between $antivirals$ and $antibiotics$. More such examples are shown in Table~\ref{tab:bad_crowd} from the Appendix. The crowd achieves overall higher annotation quality due to redundancy, over the set of 15 workers, it is unlikely they will all make the same mistake.


In Figs.~\ref{fig:cos_sim_per_worker} \& \ref{fig:ann_quality_f1} we observe that we need at least 10 workers to get a stable crowd score. This result goes against the general practice for building a ground truth, where per task there usually are 1 to 5 annotators. Based on our results, we believe that the general practice is not applicable for the use case of medical relation extraction, and should perhaps be reconsidered for other annotation use cases where ambiguity can be present, as outside of a few clear cases, the input of more annotators per task can be very useful at indicating the ambiguities inherent in language, as well as all other interpretation tasks (e.g. images, audio, event processing, etc.). Even with this added requirement, we found that crowd data is still cheaper to acquire than annotation from medical experts, as the crowd is both cheap (the cost of the crowd was $\frac{2}{3}$ that of the expert) and always available via dedicated crowdsourcing platforms like CrowdFlower.

A bottleneck in this analysis is the availability of expert annotations -- we did not have the resources to collect a larger expert dataset, and this indeed is the main reason to consider crowdsourcing. In this context, the real value of distant supervision is that large amounts of data can be gathered rather easily and cheaply, since humans are not involved. Therefore, the goal of the second experiment was to explore the trade-off between quality and cost of crowdsourcing compared to distant supervision, while scaling up the model to reach its maximum performance.

\subsection{CrowdTruth vs. distant supervision}

The results for both relations (Fig.\ref{fig:cause_3800} \& Fig.\ref{fig:treat_3800}) show that the crowd does out-perform the distant supervision baseline after the learning curves have stabilized, thus justifying its cost. From this we infer that not only is the crowd generating higher quality data than the automated baseline, but training the model with weights, as opposed to binary labels, does have a positive impact on the performance of the model.

The results of the CrowdTruth weighted F1' consistently scored above the simple F1, for both baseline and crowd over both relations.  This consolidates our assumption that ambiguity does have an impact on classifier performance, and weighting test data with ambiguity can account for this hidden variable in the evaluation.

The only weighted metric without a score increase is the baseline R' for the $cause$ relation (see Tab.\ref{tab:statistic_bas}). Recall is also the only un-weighted metric for which the $cause$ baseline model performed better than the crowd. Recall is inversely proportional to the number of false negatives, indicating that distant supervision, for this relation, is finding more positives at the expense of incorrectly labeling some of them.  This appears to be a consequence of how the model performs its training -- one of the features it learns is the UMLS type of the terms. For the $cause$ relation, it seems that term types are often enough to accurately classify a positive example (e.g. an anatomical component will rarely be the effect of a causal relation).

Over-fitting on term types classification could also be the reason that baseline performs better than the crowd in the initial experiment for $treat$ (Tab.\ref{tab:statistic_exp}), where recall for baseline is unusually high. $treat$ is also a relation that appears to favor a high recall approach -- there are very few negative examples where the type constraint of the terms (drug - disease) is satisfied. In previous work~\cite{aroyo2014threesides} we observed that $treat$ generates less ambiguity than $cause$, which explains why $treat$ has overall higher F1 scores than $cause$ in all datasets. However, the high F1 scores could also make the models for $treat$ more sensitive to confusion from ambiguous examples, as a small number of confusing sentences would be enough to decrease such a high performance. Indeed, as more (potentially ambiguous) examples appear in the training set, both the F1 and the recall of the baseline for $treat$ drop, while the crowd scores remain consistent (Fig.\ref{fig:treat_3800}). This result emphasizes the importance of weighting training data with ambiguity, as a few ambiguous examples seem to have a strong impact in generating false negatives during classification.


Our experiment has two limitations: (1) because of the limited availability of domain experts, we could not collect more than one expert judgment per sentence, and (2) because the model used classifies data with either a positive or a negative label, we removed the examples from the evaluation set that could not fit into either label.  We expect that adding more expert annotators per sentence will result in better quality annotations. However, disagreement will likely still be present -- as indicated by our previous work~\cite{aroyo2013crowd} on a set of 90 sentences, two experts agreed only 30\% of the time over what the correct relation is. Future work could explore whether disagreement between experts is consistent with the crowd disagreement. The second limitation lies with evaluation measures such as precision and recall that require discrete labels, which are the standard for classification models. The CrowdTruth method was designed specifically to represent ambiguous cases that are more difficult to fit into a positive or negative label, but to evaluate it in comparison with discrete data, we had to use the standard metrics. Now that we have shown the quality of the crowd data, it can be used to perform more detailed evaluations that take ambiguity into account through the use of weighted precision, recall and F1.

\section{Conclusion}

The standard data labeling practice used in supervised machine learning attempts to minimize disagreement between annotators, and therefore fails to model the ambiguity inherent in language. We propose the CrowdTruth method for collecting ground truth through crowdsourcing, that reconsiders the role of people in machine learning based on the observation that disagreement between annotators can signal ambiguity in the text.

In this work, we used CrowdTruth to build a gold standard of 3,984 sentences for medical relation extraction, focusing on the $cause$ and $treat$ relations, and used the crowd data to train a classification model. We have shown that, with the processing of ambiguity, the crowd performs just as well as medical experts in terms of the quality and efficacy of annotations, while being cheaper and more readily available. In addition, our results show that, when the model reaches maximum performance after training, the crowd also performs better than distant supervision. Finally, we introduced and validated new weighted measures for precision, recall, and F-measure, that account for ambiguity in both human and machine performance on this task. These results encourage us to continue our experiments by replicating this methodology for an increasing set of relations in the medical domain.

\section*{Acknowledgments}

The authors would like to thank Dr. Chang Wang for support with using the medical relation extraction classifier, and Anthony Levas for help with collecting the expert annotations.  The authors, Dr. Wang and Mr. Levas were all employees of IBM Research when the expert data collection was performed, and we are grateful to IBM for making the data freely available subsequently.

\bibliography{biblio}
\bibliographystyle{ACM-Reference-Format-Journals}

\appendix

\section{Example Sentences from the Evaluation Set}

\begin{table}[hbt!]
\centering
\caption {Example sentences removed from the evaluation (term pairs in bold font).}
\label{tab:bad_sent}
\scalebox{0.85}{
\bgroup
\def\arraystretch{1.5}
\begin{tabular}{|p{10cm}|c|>{\centering\arraybackslash}p{1.7cm}|>{\centering\arraybackslash}p{1.3cm}|}
\hline
{\bf Sentence} & {\bf Relation} & {\bf Crowd label} & {\bf Expert label} \\ \hline  \hline
The physician should ask about a history of {\bf diabetes} of long duration, including other manifestations of {\bf diabetic neuropathy}. & $cause$ & 0.977 & -1 \\ \hline
If the oxygen is too low, the incidence of {\bf decompression sickness} increases; if the {\bf oxygen} is too high, oxygen poisoning becomes a problem. & $cause$ & 0.743 & -1 \\ \hline
Evidence: ? Vigilant intraoperative magement of {\bf hypertension} is essential during resection of {\bf pherochromocytoma}. & $cause$ & -0.651 & 1 \\ \hline
This is the first case of {\bf Aicardi Syndrome} associated with lipoma and {\bf metastatic angiosarcoma}. & $cause$ & -0.909 & 1 \\ \hline
Will giving {\bf Acetaminophen} prevent the {\bf pain} of the immunization? & $treat$ & 0.995 & -1 \\ \hline
FDA approves {\bf Thalidomide} for {\bf Hansen's disease} side effect, imposes unprecedented restrictions on distribution. & $treat$ & 0.913 & -1 \\ \hline
\end{tabular}
\egroup
}
\end{table}

\begin{table}[hbt!]
\centering
\caption {Example sentences where the expert was wrong (term pairs in bold font).}
\label{tab:bad_expert}
\scalebox{0.85}{
\bgroup
\def\arraystretch{1.5}
\begin{tabular}{|p{10cm}|c|>{\centering\arraybackslash}p{1.7cm}|>{\centering\arraybackslash}p{1.3cm}|}
\hline
{\bf Sentence} & {\bf Relation} & {\bf Crowd label} & {\bf Expert label} \\ \hline  \hline
Patients with a history of {\bf bee sting allergy} may have a higher risk of a {\bf hypersensitivity reaction} with paclitaxel treatment. & $cause$ & 0.9 & -1 \\ \hline
In contrast, we did not find a definite increase in the LGL percentage within 6 months postpartum in patients with {\bf Grave's disease} who relapsed into {\bf Grave's thyrotoxicosis}. & $cause$ & 0.737 & -1 \\ \hline
{\bf Hepatoma} in one patient was correctly identified by both methods, as well as the presence of {\bf ascites}. & $cause$ & -0.579 & 1 \\ \hline
The diagnosis of {\bf gyrate atrophy} was confirmed biochemically and clinically; {\bf hyperornithinemia} and a deficiency of ornithine ketoacid transamise were confirmed biochemically. & $cause$ & -0.863 & 1 \\ \hline
Thirdly the evidence of the efficacy of {\bf Clomipramine} in {\bf OCD without concomitant depression} reported by Montgomery 1980 and supported by other studies suggests that 5 HT uptake inhibitors have a specifically anti obsessiol effect. & $treat$ & 0.905 & -1 \\ \hline
The 1 placebo controlled trial that found black cohosh to be effective for {\bf hot flashes} did not find {\bf estrogen} to be effective, which casts doubt on the study's validity. & $treat$ & 0.73 & -1 \\ \hline
{\bf Graft Versus Host Disease (GVHD) Prophylaxis} was methotrexate (1 patient), cyclosporine (2 patients), methotrexate + {\bf cyclosporine} (3 patients), cyclosporine + physical removal of T cells (2 patients). & $treat$ & -0.657 & 1 \\ \hline
Patients with severe forms of {\bf Von Willebrands' Disease (VWD)} may have frequent haemarthroses, especially when {\bf Factor VIII (FVIII)} levels are below 10 U/dL, so that some of them develop target joints like patients with severe haemophilia A. & $treat$ & -1 & 1 \\ \hline
\end{tabular}
\egroup
}
\end{table}

\begin{table}[hbt!]
\centering
\caption {Example sentences where the crowd was wrong (term pairs in bold font).}
\label{tab:bad_crowd}
\scalebox{0.85}{
\bgroup
\def\arraystretch{1.5}
\begin{tabular}{|p{10cm}|c|>{\centering\arraybackslash}p{1.7cm}|>{\centering\arraybackslash}p{1.3cm}|}
\hline
{\bf Sentence} & {\bf Relation} & {\bf Crowd label} & {\bf Expert label} \\ \hline  \hline
Instability of {\bf bone} fragments is regarded as the most important factor in pathogenesis of {\bf pseudoarthrosis}. & $cause$ & 0.928 & -1 \\ \hline
{\bf Atopic conditions} include allergic rhinitis, atopic eczema, {\bf allergic conjunctivitis} and asthma. & $cause$ & 0.507 & -1 \\ \hline
The histological finding of {\bf Psammoma bodies} is important in the diagnosis of {\bf duodel stomatostatinomas}. & $cause$ & -0.558 & 1 \\ \hline
A retrospective review of 64 patients with haematuria and subsequent histologically proven {\bf carcinoma of the bladder} revealed that bladder tumours could be diagnosed pre operatively in 34 of 46 (76\%) of patients with gross {\bf haematuria} and 12 of of 18 (67\%) of those with microhaematuria. & $cause$ & -0.658 & 1 \\ \hline
Hypersecretion of insulin increases the chance of the incidence of {\bf diabetes type I and II} while inhibiting {\bf insulin} secretion helps prevent diabetes. & $treat$ & 0.949 & -1 \\ \hline
To determine whether late asthmatic reactions and the associated increase in airway responsiveness induced by toluene diisocyate (TDI) are linked to {\bf airway inflammation} we investigated whether they are inhibited by {\bf Prednisone}. & $treat$ & 0.52 & -1 \\ \hline
In one group of four pigs sensitive to {\bf Malignant Hyperthermia (MHS)} a dose response to {\bf intravenous Dantrolene} was determined by quantitation of toe twitch tension.. & $treat$ & -0.575 & 1 \\ \hline
Deficiency diseases include night blindness and keratomalacia (caused by lack of vitamin A); beriberi and polyneuritis (lack of thiamine); {\bf pellagra} (lack of {\bf niacin}); scurvy (lack of vitamin C); rickets and osteomalacia (lack of vitamin D); pernicious anemia (lack of gastric intrinsic factor and vitamin B 12. & $treat$ & -1 & 1 \\ \hline
\end{tabular}
\egroup
}
\end{table}

\end{document}